# 基于层级化数据记忆池的边缘侧半监督持续学习方法

王祥炜　韩　锐　刘　驰

北京理工大学计算机学院　北京　100081

**摘　要**　外界环境的不断变化导致基于传统深度学习方法的神经网络性能遭受到了不同程度的下降，因此持续学习技术逐渐受到了越来越多研究人员的关注。在边缘侧环境下，面向边缘智能的持续学习模型不仅需要克服灾难性遗忘问题，还需要面对资源严重受限这一巨大挑战。这一挑战主要体现在两个方面：（1）难以在短时间内花费较大的人工开销进行样本标注，导致有标注样本资源不足；（2）难以在边缘平台部署大量高算力设备，导致设备资源十分有限。然而，面对这些挑战，一方面，现有经典的持续学习方法通常需要依赖于大量有标注样本才能维护模型的可塑性与稳定性，标注资源的缺乏将导致其准确率下降明显；但另一方面，为了应对标注资源不足的问题，半监督学习方法为了达到更高的模型准确率，往往需要付出较大的计算开销。针对这些问题，文中旨在提出一个面向边缘侧的，能够有效利用大量无标注样本及少量有标注样本的低开销的半监督持续学习方法 Edge Hierarchical Memory Learner（EdgeHML）。EdgeHML 通过构建层级化数据记忆池，使用多层存储结构对学习过程中的样本进行分级保存及回放，以在线与离线相结合的策略实现不同层级间的交互，帮助模型在半监督持续学习环境下学习新知识的同时更有效地回忆旧知识。同时为了进一步降低针对无标注样本的计算开销，EdgeHML 在记忆池的基础上，引入了渐进式学习的方法，通过控制模型对无标注样本的学习过程来减少无标注样本迭代周期。实验结果表明，在 CIFAR-10、CIFAR-100 以及 TinyImageNet 三种不同规模的数据集构建的半监督持续学习任务上，EdgeHML 相比经典的持续学习方法，在标注资源严重受限的条件下最高提升约 16.35%的模型准确率；相比半监督持续学习方法，在保证模型性能的条件下最高减少超过 50%的训练迭代时间，实现边缘侧高性能、低开销的半监督持续学习过程。

**关键词**：边缘智能；持续学习；半监督学习；数据标注；深度神经网络

# Hierarchical Memory Pool Based Edge Semi-Supervised Continual Learning Method

Xiangwei Wang, Rui Han and Chi Harold Liu

School of Computer Science and Technology, Beijing Institute of Technology, Beijing 100081, China

**Abstract**　The continuous changes in the world have resulted in the performance regression of neural networks. Therefore, continual learning (CL) area gradually attracts the attention of more researchers. For edge intelligence, the CL model not only needs to overcome catastrophic forgetting, but also needs to face the huge challenge of severely limited resources: the lack of labeled resources and powerful devices. However, the existing classic CL methods usually rely on a large number of labeled samples to maintain the plasticity and stability, and the semi-supervised learning methods often need to pay a large computational and memory overhead for higher accuracy. In response to these problems, a low-cost semi-supervised CL method named Edge Hierarchical Memory Learner (EdgeHML) will be proposed. EdgeHML can effectively utilize a large number of unlabeled samples and a small number of labeled samples. It is based on a hierarchical memory pool, leverage multi-level storage structure to store and replay samples. EdgeHML implements the interaction between different levels through a combination of online and offline strategies. In addition, in order to further reduce the computational overhead for unlabeled samples, EdgeHML leverages a progressive learning method. It reduces the computation cycles of unlabeled samples by controlling the learning process. The experimental results show that on three semi-supervised CL tasks, EdgeHML can improve the model accuracy by up to 16.35% compared with the classic CL method, and the training iterations time can be reduced by more than 50% compared with semi-supervised methods. EdgeHML achieves a semi-supervised CL process with high performance and low overhead for edge intelligence.

**Keywords**　Edge intelligence, Continual learning, Semi-supervised learning, Data label, Deep neural network

# 1 引言

随着边缘计算的不断发展，边缘智能逐渐受到了越来越多领域研究人员的关注[1-3]。例如，自动驾驶技术研究人员希望能够在边缘平台将行车过程中收集到的大量数据输入到深度神经网络（Deep Neural Network，DNN）中进行学习[4]，以提升DNN在执行图像分类等任务时的性能表现，帮助自动驾驶系统更有效地通过视觉图像对道路、行人、车辆以及基础设施等目标进行识别。然而，由于边缘平台所处地理位置的不同、时间的推移、天气的变换以及图像采集设备的规格差异等因素的影响，边缘智能系统所接收到的输入数据分布通常会随之产生变化。在典型的场景下，由于我们无法预估 DNN 模型未来需要面对的所有场景，因此需要使得 DNN 能够对新出现的（即过去不需识别甚至未曾见过的）物体进行学习，以在新的任务阶段能够对这些新的目标进行准确识别。与此同时，过往学习的物体同样可能在未来继续出现，因此我们需要使得模型在学习新知识的同时，能够尽量避免对旧知识的遗忘（即灾难性遗忘问题）[5]，权衡 DNN 的可塑性与稳定性。因此，一些研究人员提出了持续学习（Continual Learning，CL）技术予以应对。例如，研究人员通过在内存中维护一系列过去收集的数据样本，在 DNN 学习新知识的同时进行"回放"，以此维护 DNN 对旧知识的记忆。但在边缘侧环境下，现有经典的持续学习方法将受到资源严重受限这一巨大挑战。

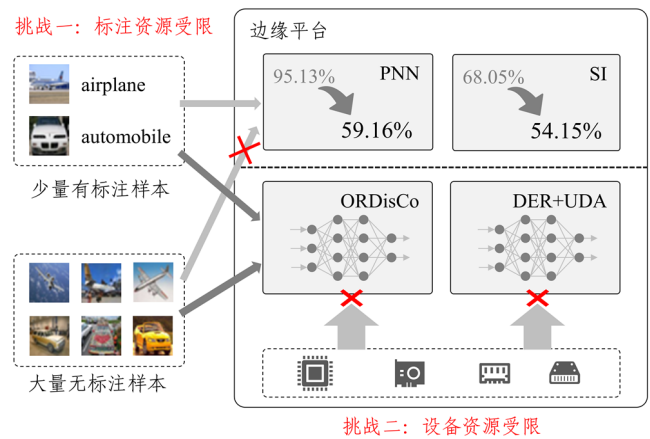

图 1 边缘侧持续学习面临的挑战
Fig. 1 The challenges of edge continual learning

如图 1 所示，资源受限带来的挑战主要体现在两个方面：（1）标注资源受限：近年来深度学习技术的快速进步以及深度神经网络模型学习效果的快速提升，离不开海量数据采集以及数据标注技术的支撑，包括图像分类模型在内的 DNN 模型通常以有监督的方式对大规模高质量标注样本进行学习[6]。但在边缘侧环境下，面对不断变化的外界环境，边缘智能系统难以在短时间内对采集到的数据进行高质量地人工标注，这种情况在输入数据涉及一些专业领域（例如医学图像）知识时将进一步恶化，因此存在着大量无法被专家标注的数据样本，导致有监督标注资源严重不足；（2）设备资源受限：边缘计算节点通常分布在广阔的网络边缘，其自身资源（例如计算资源以及内存资源）十分有限[7]，无法像数据中心一样布置大量的高算力设备（例如高性能 GPU），进而无法满足 DNN 模型长周期的迭代学习过程。

针对以上边缘智能所面临的挑战，现有的持续学习方法（例如 PNN[8]、SI[9]）通常专注于利用有标注数据来缓解灾难性遗忘问题，而没有充分挖掘无标注数据的价值，导致在标注资源受限的边缘侧环境下准确率下降明显，例如 PNN 方法准确率下降超 30%，SI 方法下降 10%以上（结果来自于前人工作[10]以及本文第 4 节）。然而，为了在持续学习过程中利用大量无标注样本，最近出现了一些关于半监督持续学习的研究工作[11]，但其为了维持模型的学习性能而引入了过多的计算及内存开销，需要保存和维护额外的生成对抗神经网络（Generative Adversarial Network，GAN）辅助分类器网络的学习。此外，将持续学习方法与典型的半监督学习方法相结合后（例如 DER[10]+UDA[12]），往往需要较长的训练时间来帮助 DNN 模型学习样本内在的知识，这对于处理器、内存等资源严重受限的边缘设备而言通常难以承受。

本文面向边缘侧半监督持续学习场景，旨在提出一种能够有效利用大量无标注样本及少量有标注样本的低开销的半监督持续学习方法 **E**dge **H**ierarchical **M**emory **L**earner（EdgeHML）。为了更有效地利用少量有标注数据以及大量无标注数据来提升 DNN 在面对环境变化时的性能表现，EdgeHML 构建了一种层级化数据记忆池，将不同类型的样本以分级管理的方式存储在不同的介质上，并在外界环境发生变化前，以离线的方式快速更新不同层级维护的样本，在模型学习的过程中在线地将内存中的有限数据进行回放。利用该层级化数据池，EdgeHML 能够在有监督持续学习的基础上，有效地挖掘大量无标注数据的内在价值。更进一步地，由于边缘设备难以像云平台一样借助大量资源进行长周期的迭代学习，EdgeHML 在层级化数据记忆池的基础上，以渐进的方式引入对无标注样本的学习过程，在保证模型性能的前提下，减少对无标注样本的迭代周期。

通过实验，EdgeHML 在 CIFAR-10、CIFAR-100 以及 TinyImageNet 三种不同规模的数据集构建的半监督持续学习任务上与基准方法进行了准确率以及计算开销两个维度的对比。结果显示，相比经典的持续学习方法，在无需引入额外 DNN 模型辅助的前提下，EdgeHML 能够在标注资源严重受限的条件下最高提升约 16.35%的准确率；同时，相比半监督持续学习方法，EdgeHML 最高减少超过 50%的训练迭代时间，有效降低了无标注样本的计算开销。

## 2 相关工作

近年来，随着深度学习技术的不断发展，持续学习领域也受到了越来越多研究人员的关注。

### 2.1 持续学习

**现有研究工作**。DNN 模型面对外界环境（包括光线、天气以及目标类别等）的不断变化，难以在认知新环境、学习新知识的同时很好地保留对旧环境中旧知识的记忆，因而造成了灾难性遗忘问题。针对这一问题，研究人员主要提出了三种思路：（1）基于正则化的思路：当模型在新任务上学习新领域的知识时，对模型的全部或部分参数加以约束，以使其能够按照特定的方向或者限制在一定的范围内进行更新，避免参数朝着错误的方向前进或偏离程度过高，造成对旧知识的遗忘。基于这一思路，SI 方法[9]受到生物学的启发，引入"智能突触"的理念，通过在线地估计损失函数对于每个参数的敏感性来计算参数重要性，并对更加重要的参数在梯度更新时施加惩罚。EWC 算法[5]同样延续正则化惩罚的思路，通过费雪信息矩阵的对角线值来估计每个参数的重要性，并将该重要性指标用于调整每个参数在惩罚项中的约束程度，以此保证模型调优的过程中不会与之前任务训练得到的模型参数产生过大的偏离。类似地，还有 MARS[13]、BLIP[14]、AGS-CL[15]、RKR[16]、VAR-GPs[17]等方法通过不同的角度，对正则化约束规则进行设计，以优化模型更新过程。（2）基于记忆的思路：将旧任务上的部分数据样本或数据特征予以保存，在新任务上学习新的数据时能够进行"复习"，避免对旧知识的遗忘。ER[18]方法会在持续学习的过程中，不断地将新遇到的样本按照特定的采样方法存入内存，在之后学习新知识的同时，从内存中取出部分样本进行学习。DER[10]在 ER 方法的基础上，提出使用过去模型对图像预测出的类别概率分布来帮助当前模型进行复习，即使用"软标签"而非类别标注这一"硬标签"使得模型能够更加充分地学习到过去的知识，从而缓解灾难性遗忘问题。此外，还有 GPM[19]、ASER[20]、MIR[21]、GSS[22]、Coresets[23]、Rainbow[24]等方法通过增强记忆样本多样性、设计数据存储的结构、衡量不同样本的贡献以及构造新的回放策略等角度来提升记忆数据对当前模型的帮助。（3）基于动态结构的思路：该类方法提出使用不同的网络部分（例如不同的参数）来学习不同的知识。例如，在遇到新任务上的新数据时，将原有网络结构进行扩展或延伸，在保留原有参数的条件下，使用新的结构部分进行学习，保障原有知识不被覆盖。又或者在一个完整的网络结构中，通过一定的策略对每个任务选取不同的参数用于学习，不同任务对应的参数之间不会产生较大的干扰，以保证对新知识和旧知识的共存。基于此思路，PNN[8]会在遇到新环境的任务时扩展 DNN 模型结构，并将原有结构与新的结构部分建立联系。在遇到需要学习或推理的样本时，利用原有结构与新添加的结构，能够使得新环境下学到的知识存储在新的结构部分中，减轻对原有知识的影响。类似地，CN-DPM[25]、PCL[26]、EFTs[27]、Kernel[28]、MERLIN[29]、Adam-NSCL[30]等方法通过设计不同的扩展结构、构造 DNN 模型参数分区策略、学习数据分布从而设计模型组装机制等维度增强基于结构的持续学习表现。

**存在的局限性**。包括上述方法在内的经典的持续学习方法通常面向有监督学习范式而设计，即输入数据均带有标注。在边缘侧环境下，由于标注资源有限，无法在短时间内获取充足的高质量人工标注，因此这些方法会在有标注样本不足的环境中出现明显的性能下降。

与有监督持续学习相反，一些研究人员更加关注于完全无监督条件下的持续学习[31,32]，但一方面，这些方法的无监督学习过程只能够对表征模型（即 DNN 网络骨干部分）进行更新，若需要真正应用于分类任务，还需额外引入 KNN 等分类器的辅助[33]；另一方面，由于缺乏对有标注样本的学习，无监督方法的性能会在较大程度上受到持续学习任务规模的影响。

### 2.2 半监督持续学习

**现有研究工作**。半监督持续学习这一场景对于 DNN 模型而言十分具有挑战性，为此 ORDisCo[11]提出具有辨别器一致性的深度在线回放方法，相互依赖地训练一个分类器和一个条件性 GAN 模型，GAN 模型会将学习到的数据分布持续地传递给分类器。ORDisCo 会以在线的方式从生成器中采样无标签数据并回放给分类器，以节省时间和空间。此外，为了克服无标签数据的灾难性遗忘，ORDisCo 会选择性地冻结辨别器中的一些参数，这些参数影响着辨别器对"无标签样本-伪标签"对的辨别。而有些研究工作为了使用少量标注样本帮助 DNN 模型快速学习新的知识，采用了元学习的方法[34]，通过预先使模型学习到一个良好的初始状态，提升模型对环境的适应能力。

**存在的局限性**。ORDisCo 等方法是通过引入额外的 GAN 模型来辅助分类器 DNN 模型对无标签数据的学习，也因此需要维护 GAN 模型的更新以及承担 GAN 模型保存、推理计算的开销，这些开销与原有分类器网络造成的开销一同施加于资源受限的边缘设备，导致其难以承受。对于基于元学习的方法而言，在外界环境变化不定的持续学习场景中，这些方法需要使用优异且具有一定规模的元训练数据集对 DNN 模型进行大量的预训练，难以适合边缘侧资源受限的场景。

综上所述，经典的持续学习方法由于难以充分利用无标注数据的价值而在边缘侧半监督持续学习场景中难以取得良好的表现，而为了挖掘大量无监督样本的价值，与半监督相结合的持续学习相关方法又往往需要引入大量额外的计算开销，例如维护额外的网络结构或模型，这使得边缘侧设备资源难以承受。针对这一问题，本文提出的 EdgeHML 能够通过层级化数据记

忆池帮助 DNN 模型从无标注样本中学习知识，并通过渐进式学习的方法减少对无标注样本的计算量，从而实现边缘侧资源高效的半监督持续学习方法。

## 3 方法

针对边缘侧持续学习方法面临的可用资源（包括标注资源与设备资源）严重受限带来的挑战，本文提出了基于层级化数据记忆池的渐进式半监督持续学习优化方法 Edge Hierarchical Memory Learner（EdgeHML），如图 2 所示，通过将特定策略下筛选出的无标注样本直接追加至硬盘池，避免对内存池的过多占用；同时，根据学习过程中不同类别样本规模以及不同类别对模型的影响值这两个维度，在每一个新任务开始前对硬盘池的样本进行采样，辅助有监督学习过程。另一方面，为了进一步降低对于无标注样本的计算开销，EdgeHML 引入了渐进式学习的方式，通过控制 DNN 模型对无监督样本的学习过程来使得模型能够在更短周期内更有效地提升性能表现。

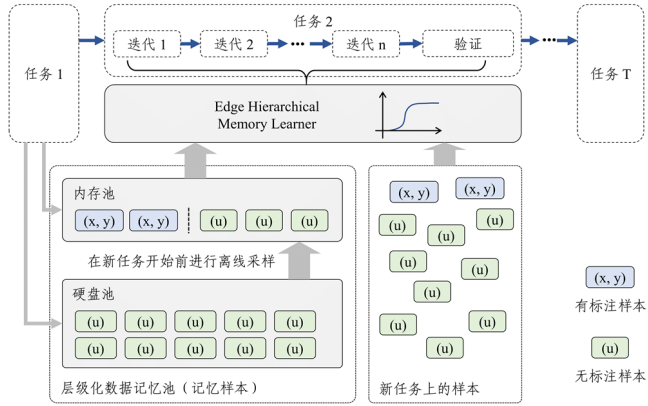

图 2 EdgeHML 总体框架图
Fig. 2 The framework of EdgeHML

**问题建模**。在基于任务增量的持续学习过程中，DNN 模型需要学习一系列的任务 $t \in \{1,2,...,T\}$，每个任务包含不同的目标物体类别。每个任务 $t$ 上包含数据 $D^t = \mathcal{X} \cup \mathcal{U}$，其中 $\mathcal{X} = \{(x_i, y_i): i \in (1,2,...,N_s)\}$ 为少量有标注样本，$\mathcal{U} = \{(u_i): i \in (1,2,...,N_u)\}$ 为大量无标注样本，其中 $N_s$ 为有标注样本数量，$N_u$ 为无标注样本数量。

### 3.1 层级化数据记忆池

EdgeHML 为了在保存无标注样本的同时不增加过多内存开销，面向边缘侧设计并实现了层级化数据记忆池（Hierarchical Memory Pool，HMP），包含内存层以及硬盘层两个层级。前者访问速度更快，但在边缘设备中容量受限；后者访问速度稍慢，但由于其廉价的特性能够在容量上远高于内存。为了弥补硬盘在 I/O 速率上的缺陷，我们针对不同类型的数据样本使用差异化的存储及回放策略，以在线与离线相结合的方式有效地协同内存层与硬盘层进行运作。

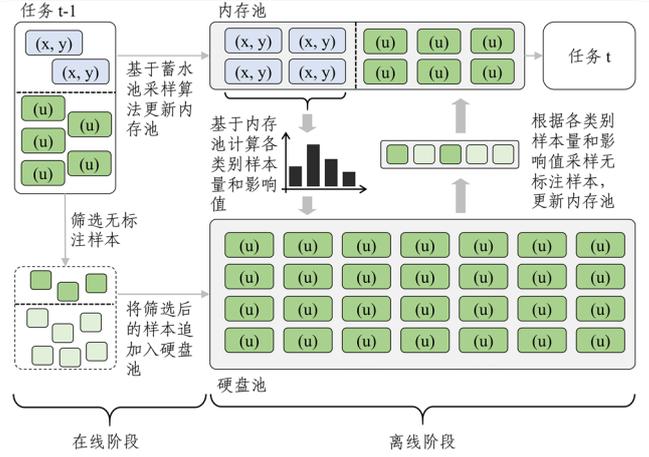

图 3 层级化数据记忆池结构图
Fig. 3 The structure of hierarchical memory pool

#### 3.1.1 硬盘层数据记忆池（硬盘池）

考虑到无标注样本的规模远高于有标注样本，因此我们在传统的基于内存的数据记忆池的基础上，设计了硬盘层数据记忆池（即硬盘池，Disk-Level Pool），硬盘池用于在持续学习过程中对特定的无标注样本进行保存。

**算法 1 硬盘池采样算法**

输入：内存池 $P_m$，硬盘池 $P_d$，总类别数 $C$，硬盘池伪标签分布 class_num []

1. 初始化类别损失统计数组 class_loss []
2. 初始化类别采样概率数组 class_prob[]
3. samples ← 从 $P_m$ 中读取有标注样本 $\{(x_i, y_i)\}$
4. for $(x_i, y_i)$ in samples do
5.     output ← $f_\Theta(x_i)$
6.     loss ← CE(output, $y_i$)
7.     class_loss[$y_i$] ← class_loss[$y_i$] + loss
8. end for
9. for i ← 1 to $C$ do
10.     class_prob[i] ← sum(class_num) / class_num[i] * class_loss[i] / sum(class_loss)
11. end for
12. class_prob ← normalize(class_prob)
13. selected_samples ← 根据 class_prob 对 $P_d$ 采样
14. 将 selected_samples 存入 $P_m$

**在线阶段**。如图 3 所示，DNN模型会在一系列的任务上遇到大量的无标注样本，我们首先使用 DNN 模型对接收到的无标注样本进行推理计算，获取模型对样本 $u_i$ 在所有类别 $c \in \{1,...,C\}$ 上的预测概率分布 $p_i^c$，当结果中最高概率值高于置信阈值 $\tau$（即 $\max(p_i^c) \geq \tau$）且对应类别 $\hat{y}_i = \underset{c}{\mathrm{argmax}}(p_i^c)$ 属于当前任务上目标类别时，将 $\hat{y}_i$ 作为该样本的伪标签，并将样本 $(u_i, \hat{y}_i)$ 作为硬盘池的候选样本。其后，针对每一个候选样本，我们进行随机采样，以一定的概率（例如 50%）决定是否采纳该候选样本作为记忆样本，并在确定采纳后将样本追加至硬盘池的尾部。同时，我们会在内存中维护对硬盘池中样本的索引，以及硬盘池中各类别（伪标签）的数量分布。该索引由于只需记录极少量的样本信息，因此不会造成过多内存开销。

考虑到硬盘的低廉成本，硬盘池的容量通常可以比内存池容量高一个数量级以上。

**离线阶段**。在每一个阶段的任务学习完成后，即将迎来下一阶段的新任务之前，EdgeHML 将在内存池与硬盘池之间进行一些离线的交互，如算法 1 所示：（1）我们根据学习过程中建立的索引对硬盘池中样本的类别进行统计，获取不同类别的数量分布，作为算法输入之一；（2）使用最新的 DNN 模型在内存池中的有标注样本上进行推理，并根据有监督学习目标计算相应的交叉熵损失，对这些损失值以类别为单位进行统计，将类别粒度的损失值作为此刻不同类别样本对 DNN 模型的影响值（第 4 至 7 行）；（3）根据不同类别样本数量以及类别粒度的影响值，确定对硬盘池中带有伪标签的无标注样本的采样概率，即某类别样本数量较高时，对应类别单个样本被采样的概率将降低，且某类别影响值较高时，对应类别单个样本被采样的概率将升高（第 9 至 12 行）；（4）根据概率分布，对硬盘池中带有伪标签的样本进行采样，并将被选取到的样本存入内存池中（第 13 至 14 行），随之后的学习过程被回放。

因此，在各个任务阶段的在线执行过程中，硬盘池中的样本无需被读取到内存中，仅需在不同任务之间离线地执行上述交互过程。

#### 3.1.2 内存层数据记忆池（内存池）

在内存中，EdgeHML 构建相应的小容量数据记忆池（即内存池，Memory-Level Pool）以存放学习过程中遇到的少量有标注样本，以及从硬盘池中采样得到的无标注样本。其中，无标注样本会在离线阶段中根据当前内存池中除有标注样本以外的剩余可用容量直接从硬盘池中采样。而对于有标注样本而言，我们会在有标注样本量小于池容量时直接存入内存池（可能会造成对内存池中无标注样本的覆盖），在有标注样本量达到池容量后基于蓄水池采样算法[35]决定是否存入。蓄水池采样算法能够在未知总数据量的情况下以平等的概率采样元素，如图 3 所示，在内存池满后，我们针对每一个从硬盘池中读取到的样本，根据内存池过往遇到的所有样本量和当前样本索引来计算蓄水池采样概率，以确定其是否被存入。值得注意的是，为了更加适应边缘侧资源条件，EdgeHML 中内存池容量在整个持续学习过程中保持不变，不会随任务发展而无限增大。

**数据回放**。随着外界环境不断变化，为了帮助 DNN 模型在学习新的知识时回忆旧的知识，从而缓解灾难性遗忘问题，EdgeHML 将在 DNN 模型学习新的有标注样本 $\{(x_i, y_i)\}$ 的同时，从内存池 $P_m$ 中进行随机采样，取出相应的记忆样本 $\{(x'_i, y'_i)\}$ 及 $\{(u'_i, \hat{p}'_i)\}$ 进行学习。

具体而言，针对新的样本，有监督学习目标为：

$$\mathcal{L}_s = \text{CE}\left(f_\Theta(x_i), y_i\right) \tag{1}$$

其中，$f_\Theta$ 表示 DNN 网络模型，$\text{CE}(\cdot,\cdot)$ 表示交叉熵。而对于记忆样本的学习目标为：

$$\mathcal{L}_m = \alpha \cdot \text{CE}\left(f_\Theta(x'_i), y'_i\right) + \beta \cdot \text{CE}(f_\Theta(u'_i), \hat{p}'_i) \tag{2}$$

其中，$\alpha$ 和 $\beta$ 分别表示针对于回放的有标注样本和无标注样本的学习权重。

### 3.2 渐进式半监督持续学习

在边缘侧环境下，计算资源的受限使得整个半监督持续学习过程无法依赖高算力的 GPU 设备进行长周期的迭代，也因此面临着更严峻的挑战。

**传统的半监督学习过程**。对于 DNN 模型，相比于单纯的有监督学习，半监督学习过程更为"曲折"，这尤其表现为 DNN 模型在半监督学习过程初期，难以快速地学习到有用的知识。然而，在有监督式学习过程中，由于输入数据具有清晰的标注，能够提供给 DNN 模型有力的指引，因此模型能够更为轻松地学习到有价值的信息。

**渐进式学习过程**。EdgeHML 结合边缘侧环境以及持续学习过程的特点，将原本在每个任务上进行的半监督学习过程视为有监督学习（$\mathcal{L}_s$）以及无监督学习（$\mathcal{L}_u$）两个子过程。在持续学习过程中，每当新任务到来的初期，除数据回放学习目标外，EdgeHML 仅使用有监督学习目标来支撑 DNN 模型的学习，经过一定时间的训练后，再逐渐地引入对于无标注样本的学习目标。为了使得模型能够在外界环境发生变化后的初期不会面临难度陡增的学习目标，而是由相对简单的有标注样本学起，平缓地增加对于无标注样本的学习，并最终过渡到更为复杂的半监督学习过程中，EdgeHML 尝试利用多种函数曲线对无标注样本学习目标的权重进行控制，并经过实验评估后使用 Cosine 函数进行实现。具体而言，EdgeHML 设定阈值 $v_1$ 和 $v_2$，当在新任务上的迭代周期 $v$ 达到阈值 $v_1$ 后，基于 Cosine 函数计算无监督权重 $\gamma$，当迭代周期 $v$ 超过阈值 $v_2$ 后固定该学习目标权重，直到当前任务结束，即：

$$\gamma = \begin{cases} 0, & v < v_1 \\ \eta \cdot \cos\left(\frac{\pi \cdot (v - v_1)}{(v_2 - v_1)}\right) + \xi, & v \geq v_1 \cap v < v_2 \\ \eta \cdot \cos(\pi) + \xi, & v \geq v_2 \end{cases} \tag{3}$$

其中，$\eta$ 和 $\xi$ 用于控制无标注样本学习目标权重的变化过程，通常分别取-0.5 及 0.5 即可（该参数面对不同任务时无需特殊调整）。由于 $v < v_1$ 时，$\gamma$ 为 0，因此这段时间内无需针对无监督学习目标进行计算。其中，$v_1$ 通常取完整迭代周期数的 20%，$v_2$ 通常取完整迭代周期数的 21% 至 30% 即可，即每个任务开始的前 20% 的迭代周期内无需计算无监督学习目标对应的损失值，实现了在保证最终性能的同时，降低在计算设备上的开销。与此同时，我们也在实验部分评估了 EdgeHML 在更低资源开销（即 $v_1$ 取完整迭代周期数的 60%）条件下的准确率和计算开销。

综上所述，对于半监督持续学习中的每个任务，整体学习目标为：

$$\mathcal{L} = \mathcal{L}_s + \mathcal{L}_m + \gamma \cdot \mathcal{L}_u \quad (4)$$

其中，$\mathcal{L}_u$ 中置信阈值 $\tau$ 通常取 0.95[36]：

$$\mathcal{L}_u = \mathbb{1}(\max(p_i^c) \geq \tau) \cdot \text{CE}(f_\Theta(u_i), \hat{p}_i) \quad (5)$$

通过公式 4 所示的学习目标优化，本文所提方法 EdgeHML 能够综合利用在外界环境变化时遇到的新数据以及层级化数据记忆池中的历史数据帮助模型在学习新知识的同时，加强对旧知识的记忆，提升 DNN 模型在持续学习过程中的整体表现。

## 4 实验结果与分析

在本节中，我们使用三种不同规模的数据集，将本文所提方法 EdgeHML 与基准方法在准确率以及计算开销两方面进行实验对比，并对结果进行分析。

### 4.1 评估指标

为了更加准确地评估不同方法的性能，我们使用两方面的指标进行评估：

（1）平均准确率（Accuracy）：为了评估在标注资源受限的条件下，EdgeHML 与经典的持续学习方法的学习效果，本文使用平均准确率进行评估，即在学习完最后一个任务 $T$ 之后，在所有任务 $t \in \{1, 2, \ldots, T\}$ 上进行测试，得到准确率 $a_t^T$ 的均值，即：

$$\text{Accuracy} = \frac{1}{T}\sum_{t=1}^{T} a_t^T \quad (6)$$

（2）训练迭代时间开销（Iterations Time）及无标注样本迭代周期（Unsupervised Iterations）：为了评估 EdgeHML 与半监督持续学习方法的计算开销，尤其是对于无标注样本的计算量，本文记录不同方法在持续学习过程内的训练迭代时间（该时间结果还包含了 EdgeHML 中离线阶段的耗时，包括内存池和硬盘池的访问时间，排除了所有方法在模型验证等无关过程中的耗时），以及展示不同方法中对于无标注样本的迭代周期占总迭代周期的比例。即在总迭代周期一致的条件下，Unsupervised Iterations 占比越小，则对于无标注样本的迭代周期越少。

### 4.2 实验设置

**数据集**。根据半监督方法以及持续学习方法的相关研究工作[10,32,36,40]，我们采用不同规模的 CIFAR-10 数据集[37]、CIFAR-100 数据集[37]以及 TinyImageNet 数据集[38]构造半监督持续学习任务，持续学习的设定为任务增量。具体包括：（1）CIFAR-10：完整的 CIFAR-10 数据集包括飞机、汽车、鸟、猫等 10 个物体类别，我们将 CIFAR-10 按照类别划分为 5 个任务，即每个任务包括 2 个类别，不同任务所包含的类别互不重叠。在每个任务的每个类别上，我们仅使用 5 个样本及对应的标注作为有标注数据，而包括这 5 个样本在内的所有其余样本将作为无标注数据，构建半监督持续学习任务 Split Semi CIFAR-10-5；（2）CIFAR-100：CIFAR-100 数据集包含 100 个物体类别，我们按照类别将其划分为 20 个任务，每个任务包含 5 个不同的类别。类似地，我们在每个类别上使用 5 个样本作为有标注样本，其余所有样本（包含这 5 个）作为无标注样本，构建半监督持续学习任务 Split Semi CIFAR-100-5；（3）TinyImageNet：TinyImageNet 数据集是 ImageNet 数据集[39]的变体，包含 200 个物体类别，根据前人研究工作[40]，本文将 TinyImageNet 数据集中的图像缩放到 64*64 像素，按照类别划分为 10 个任务，每个任务包含 20 个类别。每个类别的样本中包含 5 个有标注样本作为有标注数据，包含这 5 个样本的所有其余样本作为无标注数据，构建半监督持续学习任务 Split Semi TinyImageNet-5。

**基准方法**。我们从朴素训练方法、持续学习方法以及半监督持续学习方法这三个维度选取基准方法，以面向边缘侧受限环境条件进行对比实验验证，具体包括：（1）朴素方法（Supervised Fine-Tune，SFT）：面对每一个新的任务，朴素方法直接使用 DNN 模型在少量有标注样本上进行调优；（2）持续学习方法：我们分别按照持续学习的三种主流思路选取了典型的方法进行对比，即基于正则化约束的方法 SI[9]、基于结构的方法 PNN[8]以及基于内存回放的方法 DER[10]；（3）半监督持续学习方法：我们选取了半监督持续学习方法 ORDisCo[11]、无监督持续学习方法 LUMP[32]，并将现有的持续学习方法与典型的半监督学习方法相结合，设置了 DER + FlexMatch[41]、DER + UDA[12]作为基准方法。其中，所有需要数据回放的方法（除特殊说明外）均采用统一的内存池容量。

**网络结构及训练**。基于前人研究工作[10,32]中的实验设置，为了更加合理地验证不同方法在边缘侧环境下的性能，我们使用深度神经网络结构 ResNet-18[42]进行不同方法的实现，且不使用预训练对 DNN 模型权重进行初始化。训练过程中学习率设为 0.03[10,32]，EdgeHML 中的 $\alpha$ 设为 1，$\beta$ 设为 0.1。

**实验环境**。我们使用 NVIDIA 面向边缘推出的 Jetson AGX Orin 平台进行实验，该平台采用 ARM 架构，具有 8 核 CPU 以及 32 GB 内存。

### 4.3 结果及分析

#### 4.3.1 准确率结果及分析

表 1  不同方法的准确率对比结果
Table 1  Accuracy of each method

| 方法 | Split Semi CIFAR-10-5 / % | Split Semi CIFAR-100-5 / % | Split Semi TinyImageNet-5 / % |
| --- | --- | --- | --- |
| SFT | 55.48 | 24.37 | 8.66 |
| SI | 54.15 | 30.80 | 9.84 |
| PNN | 59.16 | 33.71 | 12.51 |
| DER | 62.15 | 36.24 | 15.83 |
| ORDisCo | 65.91 | - | - |
| LUMP (SimSiam) | 63.11 | 39.32 | 13.00 |
| DER+FlexMatch | 61.88 | 40.48 | 16.10 |
| DER+UDA | 64.50 | 40.20 | 17.10 |
| **EdgeHML (ours)** | **66.83** | **52.59** | **22.55** |

**现有方法**。如表 1 所列，经典的持续学习方法在面对有标注样本数量较少的情况时表现不佳，尤其是基于正则化以及基于结构的持续学习方法 SI 和 PNN，在 Split Semi CIFAR-10-5 任务上准确率难以超过 60%，在数据量更大的 Split Semi TinyImageNet-5 以及任务数更多的 Split Semi CIFAR-100-5 上的准确率分别难以超过 15%、35%，其中 SI 方法相比朴素方法 SFT 而言效果微弱，在有监督资源严重受限的情况下难以很好地缓解灾难性遗忘问题，而基于数据回放的 DER 方法则相对发挥了一定程度的作用，能够更大程度地利用少量有标注样本的价值。此外，对于无监督持续学习方法 LUMP，由于其能够利用大量的无标注样本进行学习，因此在任务数量更大（即持续学习过程更久）的 Split Semi CIFAR-100-5 上准确率稍好。而对于半监督持续学习方法而言（由于 ORDisCo 方法与本文在 CIFAR-100、TinyImageNet 数据集上采用的典型场景设置[10,40]不同，因此仅展示了其在 CIFAR-10 数据集上的准确率结果），由于能够同时利用少量有标注样本以及大量无标注样本进行学习，因此最终准确率相较经典的持续学习方法而言有不同程度的提升。

**本文方法**。相比于基准方法，EdgeHML 能够通过层级化数据记忆池对有标注以及无标注样本进行有效地回放学习，在三种数据集上分别实现了 66.83%、52.59%以及 22.55%的准确率。其中，在任务数量最大的 Split Semi CIFAR-100-5 上相比经典的持续学习方法（最高值为 36.24%）提升了 16.35%，相比无监督持续学习方法（最高值为 39.32%）提升了 13.27%；在数据量最大的 Split Semi TinyImageNet-5 上相比经典的持续学习方法（最高值为 15.83%）提升了 6.72%。由此可见，EdgeHML 在更加困难的任务中能够更大程度地释放无标注样本的价值，从而获得更大的准确率提升。

与此同时，为了更进一步地对比 EdgeHML 与经典的持续学习方法在整个持续学习过程中各个任务上的表现，我们还记录了他们在学习完每个任务后，对当前及之前所有任务进行测试的平均准确率。如图 4 所示，经典的持续学习方法在第 1 个任务上能够较快地学习知识，取得更高的准确率，但随着目标类别的变化以及有标注样本数量的限制，导致其在第 2 个任务上即出现了明显的性能下降。性能下降的趋势在任务中期有所缓解，直到后期准确率逐渐上升，但受限于无法挖掘大量无监督数据的价值，因此准确率上升速度缓慢。EdgeHML 虽然在初期对 DNN 模型性能提升有限，但随着任务的变化，能够凭借对有标注数据以及无标注数据的综合学习，快速提升 DNN 模型的准确率，并在任务后期实现了高于 65%的准确率水平。

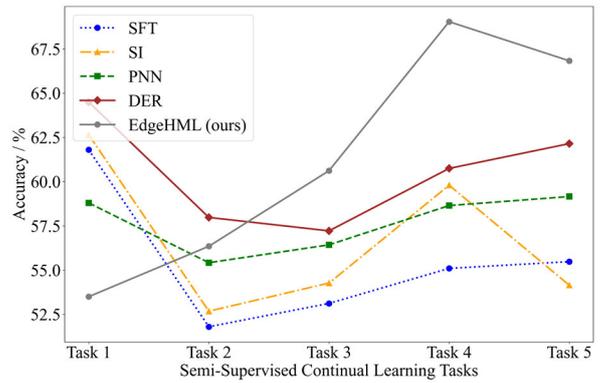

图 4  不同方法在 Split Semi CIFAR-10-5 的各个任务上的表现
Fig. 4  Task-wise performance of each method on Split Semi CIFAR-10-5

### 4.3.2 计算开销结果及分析

表 2  不同方法的计算开销对比结果
Table 2  Computation cost of each method

| 方法 | Accuracy / % | Unsupervised Iterations / % | Iterations Time / seconds |
| --- | --- | --- | --- |
| SFT | 55.48 | 0 | 129.42 |
| SI | 54.15 | 0 | 158.76 |
| PNN | 59.16 | 0 | 392.58 |
| DER | 62.15 | 0 | 241.56 |
| LUMP (SimSiam) | 63.11 | 100 | 295.38 |
| DER+FlexMatch | 61.88 | 100 | 484.56 |
| DER+UDA | 64.50 | 100 | 530.64 |
| **EdgeHML (ours)** | **66.83** | **80** | **298.74** |
| **EdgeHML-60 (ours)** | **66.09** | **40** | **229.26** |

如表 2 所列：（1）对于无标注样本迭代周期，基准方法中，经典的持续学习方法由于不参与无监督计算，因此比例为 0；无监督持续学习方法 LUMP 以及其余半监督相关方法均在全程对无标注样本进行计算，因此迭代比例为 100%。本文所提方法 EdgeHML 在层级化数据记忆池的基础上，引入了渐进式学习的方式，因此能够控制对于无标注样本的迭代周期。在表 2 中，EdgeHML 分别展示了正常设置（经过 20%的总迭代周期后引入无监督计算，即此时 $v_1$ 取完整迭代周期数的 20%）以及更加偏向资源高效的设置 EdgeHML-60（经过 60%的总迭代周期后引入无监督计算，即此时 $v_1$ 取完整迭代周期数的 60%）两种设置下的准确率及开销。相应地，这两种设置中对于无标注样本的学习周期占总周期的比例分别为 80%、40%，后者由于对无标注样本的学习周期更少，因此计算开销相对更低。

（2）训练迭代时间指标通过实验测定 DNN 模型在 Split Semi CIFAR-10-5 任务上的训练迭代时间，更为直观地展现了不同方法在实际设备中的计算开销。对于经典的持续学习方法而言，只对少量有标注样本进行计算，其中，基于结构扩展的 PNN 方法由于网络结构不断延展导致计算开销逐渐增长，除此以外的经典持续学习方法整体上训练迭代时间更短，计算开销更低，但在复杂的半监督持续学习场景下，其准确率水平也相应更低。而无监督持续学习方法不会对有标注样本进行学习，因此其计算开销主要来源于针对无标注样本的学习过程，由于无标注数据本身规模较大，因此训练迭代时间相比 DER 等方

法稍高。半监督持续学习方法（例如 DER+UDA）能够综合利用有标注样本与无标注样本进行半监督学习，也因此造成了更高的计算开销，具体体现为训练迭代时间的大幅延长。

而 EdgeHML 对于不同类型数据的学习目标更为简单，并且能够利用不同层级的数据记忆池对不同样本进行分级管理，更进一步地利用渐进式学习的方法控制对于无标注样本的学习周期，因此可以在模型准确率以及计算开销二者之间进行有效权衡。具体而言，当 EdgeHML 经过 20%的迭代周期后加入无监督学习时，能够达到 66.83%的准确率，即在训练迭代时间接近 LUMP 方法的条件下，准确率相比 LUMP 提升了超过 3%；而当 EdgeHML 经过 60%的迭代周期后再开始无监督学习时（即 EdgeHML-60），训练迭代时间进一步缩短到了 229.26 秒（包含离线阶段的耗时），已经接近于经典的持续学习方法，同时相比半监督持续学习方法（例如 DER+FlexMatch 等）降低了超过 50%的时间，而准确率仍然能够保持在 66%以上。

### 4.3.3 不同标注数量与池容量条件下结果及分析

在上述实验的基础上，为了更进一步地验证 EdgeHML 在不同环境设定下的性能表现，我们额外设置了两组实验：（1）不同的标注数量：参考前人工作[11,36,41]中的实验设置，我们分别对每个类别提供 5 个、25 个以及 100 个样本作为有标注样本，其余样本作为无标注样本，构造 Split Semi CIFAR-100-5、Split Semi CIFAR-100-25 以及 Split Semi CIFAR-100-100 三个半监督持续学习任务，评估方法在不同标注资源条件下的准确率；（2）不同的池容量：不同的边缘设备可能具有不同规模的数据存储资源，针对这一点，我们参考前人工作[10]设置了三种内存池容量（即 200、500、2000 个样本）以及相应的硬盘池容量（即 10000、12000、15000 个样本），使用 Split Semi CIFAR-100-5 任务评估 EdgeHML 在不同容量的数据记忆池条件下的性能表现。

表 3  不同标注数量条件下各方法准确率结果
Table 3  Accuracy of each method with different labels

| 方法 | Split Semi CIFAR-100-5 / % | Split Semi CIFAR-100-25 / % | Split Semi CIFAR-100-100 / % |
|---|---|---|---|
| SFT | 24.37 | 27.60 | 31.45 |
| SI | 30.80 | 31.80 | 32.63 |
| PNN | 33.71 | 46.32 | 46.15 |
| DER | 36.24 | 49.62 | 51.86 |
| LUMP (SimSiam) | 39.32 | 48.36 | 60.32 |
| DER+FlexMatch | 40.48 | 50.55 | 54.65 |
| DER+UDA | 40.20 | 52.34 | 53.81 |
| **EdgeHML (ours)** | **52.59** | **68.33** | **69.60** |

**不同标注数量**。如表 3 所列，各方法在标注数量增多时准确率整体呈现出上升趋势，这表明标注资源对于持续学习方法的性能影响较大。尤其是在标注资源整体规模较小的情况下，例如每类别标注从 5 个上升至 25 个时，（包括 EdgeHML 在内的）各方法准确率提升明显。本文所提方法 EdgeHML 在每类别标注数量为 25 以及 100 时，准确率分别达到了 68.33%以及 69.60%。与基准方法相比，EdgeHML 在不同标注数量条件下均提升了超过 9%的准确率。

表 4  不同池容量条件下 EdgeHML 结果
Table 4  Performance of EdgeHML with different pool sizes

| 池容量 | Accuracy / % | Iterations Time / seconds |
|---|---|---|
| 200+10000 | 43.05 | 291.51 |
| 500+12000 | 50.48 | 336.74 |
| 2000+15000 | 52.59 | 345.99 |

**不同池容量**。如表 4 所列，一方面，随着内存池以及硬盘池容量的增长，EdgeHML 的准确率整体呈现出上升的趋势，迭代时间也随之上升。实验结果表明数据记忆池的容量能够对最终准确率产生较大影响，这一点与前人研究工作[10]类似。另一方面，虽然基于数据回放的方法的性能在一定程度上依赖于数据记忆池的规模，但 EdgeHML 在内存池容量十分有限（即仅能容纳 200 个样本）的条件下，仍然达到了 43.05%的准确率，（由表 3 可知）与基准方法相比表现更优。

**实验结果总结**。综上所述，EdgeHML 通过层级化数据记忆池能够更加有效地帮助 DNN 模型从有标注以及无标注样本中进行学习，相比经典的持续学习方法最高提升了 16.35%的准确率；同时，在对有标注样本以及无标注样本进行分级管理的基础上，EdgeHML 利用渐进式半监督持续学习的方法进一步降低了计算开销，相比半监督持续学习方法最高减少了超过 50%的训练迭代时间，实现了在边缘侧标注资源有限的条件下，准确率以及资源开销两方面的改善。

**结束语**  持续学习技术在边缘侧环境下面临着有监督标注资源不足以及设备资源不足这两方面的挑战。为了改善现有经典的持续学习方法难以充分利用无标注样本这一问题，本文提出了边缘侧半监督持续学习方法 EdgeHML，通过构建层级化的数据记忆池以及在线与离线相结合的层级间交互策略，能够对有标注样本以及无标注样本进行高效的分级管理及回放学习，利用筛选后的无标注样本帮助模型更有效地在新环境下掌握新的知识。同时，在层级化数据记忆池的基础上，EdgeHML 能够通过渐进式学习的方式减少针对无标注样本的计算量，有效降低半监督持续学习过程的训练迭代时间，在保证 DNN 模型性能的条件下进一步减轻边缘设备的资源负担。通过在不同规模的 Split Semi CIFAR-10-5、Split Semi CIFAR-100-5 以及 Split Semi TinyImageNet-5 半监督持续学习任务上的对比测试，实验结果验证了 EdgeHML 的良好性能。本文工作仍存在一定的局限性，即只专注于边缘侧任务增量型的持续学习任务。在未来的工作中，我们将进一步探索在领域增量型等持续学习任务上的数据记忆池的结构设计以及记忆样本的学习方式。